\documentclass[11pt]{article}
\usepackage[margin=1in]{geometry}
\usepackage[T1]{fontenc}
\usepackage[utf8]{inputenc}
\usepackage{lmodern}
\usepackage{microtype}
\usepackage{amsmath,amssymb}
\usepackage{booktabs}
\usepackage{graphicx}
\usepackage{subcaption}
\usepackage{caption}
\usepackage{tikz}
\usetikzlibrary{arrows.meta,positioning}
\usepackage[round]{natbib}
\usepackage{placeins}
\usepackage[hidelinks]{hyperref}
\newcommand{\NALdistrPTwo}{.192}
\newcommand{\NALmFour}{.393}
\newcommand{\NALmOne}{.857}
\newcommand{\NALmSixteen}{.182}
\newcommand{\NALpTwo}{.389}
\newcommand{\NBGchance}{.167}
\newcommand{\NBGcot}{.850}
\newcommand{\NBGdirect}{.580}
\newcommand{\NBQToD}{$-$.403}
\newcommand{\NBQToDhi}{$-$.339}
\newcommand{\NBQToDlo}{$-$.467}
\newcommand{\NBQToF}{.576}
\newcommand{\NBQToFhi}{.629}
\newcommand{\NBQToFlo}{.523}
\newcommand{\NBQToM}{.173}
\newcommand{\NBQToMhi}{.209}
\newcommand{\NBQToMlo}{.135}

\newcommand{\NBQTtD}{$-$.398}
\newcommand{\NBQTtDhi}{$-$.336}
\newcommand{\NBQTtDlo}{$-$.459}
\newcommand{\NBQTtF}{.628}
\newcommand{\NBQTtFhi}{.679}
\newcommand{\NBQTtFlo}{.576}
\newcommand{\NBQTtM}{.230}
\newcommand{\NBQTtMhi}{.274}
\newcommand{\NBQTtMlo}{.186}
\newcommand{\NBQTtMreadsF}{.352}
\newcommand{\NBQTtaftF}{.900}
\newcommand{\NBQTtbefF}{.731}
\newcommand{\NBQTthF}{.868}
\newcommand{\NBQTthM}{.324}
\newcommand{\NBQTtnonmodeM}{.626}
\newcommand{\NBQTtnonmodeMaj}{.189}
\newcommand{\NBQTtpreF}{.402}
\newcommand{\NBQTtpreFhi}{.480}
\newcommand{\NBQTtpreFlo}{.321}
\newcommand{\NBQTtwN}{242}
\newcommand{\NBQTtwrongF}{.798}
\newcommand{\NBQTtwrongGold}{.115}
\newcommand{\NBQn}{467}
\newcommand{\NBQnWrong}{102}
\newcommand{\NBQoracleM}{.577}
\newcommand{\NBQruleF}{.175}
\newcommand{\NBQruleM}{$-$.062}
\newcommand{\NBQseedDmax}{$-$.419}
\newcommand{\NBQseedDmin}{$-$.382}
\newcommand{\NBQwithinH}{1.18}
\newcommand{\NEBdf}{.615}
\newcommand{\NEBdfhi}{.667}
\newcommand{\NEBdflo}{.562}
\newcommand{\NEBdr}{.547}
\newcommand{\NEBdrhi}{.597}
\newcommand{\NEBdrlo}{.496}
\newcommand{\NEBfinNone}{.003}
\newcommand{\NEBfinSwap}{.005}

\newcommand{\NEBnorm}{.003}
\newcommand{\NEBrand}{.086}

\newcommand{\NEBsameKeepB}{.982}

\newcommand{\NEBswap}{.700}
\newcommand{\NECbsixM}{6.3M}
\newcommand{\NECcells}{24}
\newcommand{\NECchance}{.176}
\newcommand{\NECmaxdev}{.193}
\newcommand{\NECmindev}{.157}
\newcommand{\NECthr}{.226}
\newcommand{\NECtrainPthree}{.959}
\newcommand{\NECtrainPzero}{.300}
\newcommand{\NEED}{.422}
\newcommand{\NEEDhi}{.478}
\newcommand{\NEEDlo}{.364}
\newcommand{\NEEF}{.303}
\newcommand{\NEEFfourteen}{.278}
\newcommand{\NEEM}{.794}
\newcommand{\NEEMfourteen}{.836}
\newcommand{\NEEacc}{.699}
\newcommand{\NEEend}{.915}
\newcommand{\NEEgpqaF}{15}
\newcommand{\NEEgpqaN}{198}
\newcommand{\NEEmmluF}{670}
\newcommand{\NEEmmluN}{1000}
\newcommand{\NEEn}{432}
\newcommand{\NEEnullF}{.267}
\newcommand{\NEEpost}{.332}
\newcommand{\NEEqtr}{.294}
\newcommand{\NEEqtrz}{1.57}
\newcommand{\NEEsecF}{.408}
\newcommand{\NEEseczF}{7.78}
\newcommand{\NEEshuf}{.426}
\newcommand{\NEEzF}{1.69}
\newcommand{\NEFalpha}{0.4}

\newcommand{\NEFauditSteer}{20}

\newcommand{\NEFgtwo}{.000}
\newcommand{\NEFlayer}{20}
\newcommand{\NEFn}{400}
\newcommand{\NEFreplay}{.9975}
\newcommand{\NESacc}{.266}
\newcommand{\NESbal}{.265}
\newcommand{\NESchance}{.250}
\newcommand{\NESgone}{1.000}
\newcommand{\NESmaj}{.276}

\newcommand{\NESptenth}{.882}
\newcommand{\NESz}{1.26}
\newcommand{\NLDD}{.575}
\newcommand{\NLDDhi}{.632}
\newcommand{\NLDDlo}{.514}
\newcommand{\NLDF}{.240}

\newcommand{\NLDM}{.867}
\newcommand{\NLDPoneX}{.867}
\newcommand{\NLDPtenth}{.289}
\newcommand{\NLDPthirtieth}{.217}

\newcommand{\NLGchance}{.179}
\newcommand{\NLGcot}{.511}
\newcommand{\NLGdirect}{.178}

\newcommand{\NLYSixteenF}{.173}
\newcommand{\NLYSixteenM}{.204}

\newcommand{\NLYTwentyEightF}{.204}
\newcommand{\NLYTwentyEightM}{.966}

\newcommand{\NLYTwentyF}{.212}

\newcommand{\NLYTwentyFourF}{.196}
\newcommand{\NLYTwentyFourM}{.946}

\newcommand{\NLYTwentyM}{.906}

\newcommand{\NLYgzero}{.694}
\newcommand{\NLYmaxzF}{2.12}
\newcommand{\NMXNiD}{$-$.356}
\newcommand{\NMXNiDhi}{$-$.306}
\newcommand{\NMXNiDlo}{$-$.403}

\newcommand{\NMXTsD}{$-$.108}
\newcommand{\NMXTsDhi}{$-$.056}
\newcommand{\NMXTsDlo}{$-$.161}

\newcommand{\NMXTsseedOtherDa}{$-$.318}
\newcommand{\NMXTsseedOtherDb}{$-$.352}
\newcommand{\NMXTsseedTwoD}{+.345}
\newcommand{\NMXcOneHeads}{12}
\newcommand{\NMXcOneTrainOk}{2}
\newcommand{\NMXcOneValOk}{5}
\newcommand{\NMXcThreeFhi}{1.64}
\newcommand{\NMXcThreeFlo}{1.00}
\newcommand{\NMXcThreeMhi}{1.91}
\newcommand{\NMXcThreeMlo}{1.80}
\newcommand{\NMXcThreeTosFin}{1.905}
\newcommand{\NMXcThreeTosZero}{1.771}
\newcommand{\NMXcThreeTtsFin}{1.831}
\newcommand{\NMXcThreeTtsZero}{1.739}
\newcommand{\NMXcTwoHeads}{12}
\newcommand{\NMXcTwoTrainOk}{11}
\newcommand{\NMXcTwoValOk}{12}
\newcommand{\NMXconE}{$-$.008}
\newcommand{\NMXconEhi}{.052}
\newcommand{\NMXconElo}{$-$.069}
\newcommand{\NMXconL}{$-$.263}
\newcommand{\NMXconLhi}{$-$.199}
\newcommand{\NMXconLlo}{$-$.324}
\newcommand{\NMXddCE}{1.788}
\newcommand{\NMXn}{532}
\newcommand{\NMXpairN}{415}
\newcommand{\NMXtotal}{24}
\newcommand{\NMXvalid}{21}
\newcommand{\NQBD}{.511}
\newcommand{\NQBDhi}{.569}
\newcommand{\NQBDlo}{.451}
\newcommand{\NQBF}{.166}

\newcommand{\NQBM}{.705}
\newcommand{\NQBPoneX}{.879}
\newcommand{\NQBPtenth}{.177}

\newcommand{\NQBchF}{.173}

\newcommand{\NQBpriN}{61}

\newcommand{\NQDD}{.696}
\newcommand{\NQDDhi}{.744}
\newcommand{\NQDDlo}{.648}
\newcommand{\NQDF}{.221}
\newcommand{\NQDFexcess}{.035}
\newcommand{\NQDM}{.966}
\newcommand{\NQDPoneX}{.975}
\newcommand{\NQDPtenth}{.317}
\newcommand{\NQDPthirtieth}{.249}
\newcommand{\NQDPthirtiethexcess}{.064}
\newcommand{\NQDchF}{.186}
\newcommand{\NQDchM}{.234}
\newcommand{\NQDn}{353}
\newcommand{\NQDsecD}{.511}
\newcommand{\NQDsecDhi}{.549}
\newcommand{\NQDsecDlo}{.473}
\newcommand{\NQDsecF}{.232}
\newcommand{\NQDsecFexcess}{.045}
\newcommand{\NQDsecN}{883}
\newcommand{\NQDsecchF}{.187}
\newcommand{\NQDseczF}{3.45}
\newcommand{\NQDzF}{1.70}
\newcommand{\NQDzM}{32.4}
\newcommand{\NQTchance}{.180}
\newcommand{\NQTitems}{1040}
\newcommand{\NQTmid}{.181}
\newcommand{\NQTpre}{.569}
\newcommand{\NQTprompt}{.178}
\newcommand{\NQTroll}{.987}

\date{2026-11-06}   

\title{When Answer Decodability Is Not a Decision:\\
Separating Final-Answer and Imminent-Emission Signals in Chain-of-Thought}
\author{Hyojung Han$^{\dagger}$ \qquad Jongmin Kim \qquad Seung-Hun Jeon\\[4pt]
        \small ThakiCloud\\[2pt]
        \small $^{\dagger}$Corresponding author: \texttt{hyojung.han@thakicloud.com}}

\begin{document}
\maketitle

\begin{abstract}
Answer probes applied to chain-of-thought hidden states are often read as evidence that a language model
has already decided its final response. This reading overlooks a competing signal, the candidate the
model is about to mention. We introduce emission-controlled dual-target probing (ECDP), which fits the
same low-rank reader family separately to the imminent mention and to the final answer at identical
positions, with target-specific nulls and thresholds fixed before any fit. On correct DecisionDepth (DD)
rollouts in which a distractor is mentioned before the final answer, Qwen3.5-4B decodes the next mention
at \NQDM{} (null \NQDchM) but the final answer only at \NQDF{} (null \NQDchF), an emission gap of
\NQDD{} [\NQDDlo, \NQDDhi]; Llama-3.1-8B gives \NLDD{} [\NLDDlo, \NLDDhi]. These brackets are descriptive
item-bootstrap intervals from the median-gap seed among three post-hoc robustness runs; they exclude training-seed variation and are not adjusted
for multiplicity. This pattern is not universal. In a pre-specified evaluation of Qwen3-8B with thinking
enabled on held-out MMLU questions, using a full-prefix attention reader over every thinking position, both
targets are decodable and the final answer is read far more strongly (mention-minus-answer lift
\NBQTtD{} [\NBQTtDlo, \NBQTtDhi] at layer 22 over \NBQn{} questions); an exploratory, post-hoc run of
Qwen3.5-4B under the same protocol points the same way at layer 19. Model, task, positions, and reader
change together between these settings, and an attempted matched comparison on DD produced no valid
statistics because the imminent-mention probe did not train reliably, so we do not identify what reverses
the order. On DD, planted-signal controls show that the reader recovers an injected final-answer signal at
one tenth of full dose. Answer-candidate decodability is therefore target-relative and protocol-dependent: a
strong answer readout does not by itself establish when a model commits, and a probe that fails to train is
not evidence that the information is absent. ECDP is a practical control for distinguishing imminent text
emission from evidence of a settled decision.
\end{abstract}

\section{Introduction}

Hidden-state probing has become a standard way to ask when a language model ``knows'' its answer during
chain-of-thought (CoT) reasoning \citep{wei2022,kojima2022}. Recent work reports that final answers on
easy multiple-choice questions are decodable early in the rationale and uses this for early exit
\citep{theater2026}, and that probes reveal how far ahead a model represents its future reasoning
\citep{farahead2026}. A common step in these analyses is to treat a successful answer probe as evidence
of a decision state.

That step has a confound. When a rationale weighs several candidates, it writes their names. Just before
a candidate's name appears, the hidden state must carry what is needed to produce that name. An
imminent-emission signal is therefore a plausible competing source of decodability whenever the candidate
about to be emitted coincides with the target a probe is scored on. Whether it tracks the eventual choice is a separate question, and the usual evaluation does not
separate the two.

The pattern is visible even without a control. In our rollouts the final answer is at chance sixteen
tokens before it is first written in the conclusion and reaches \NALmOne{} one token before it
(Figure~\ref{fig:align}), so a high answer readout can arise exactly when the answer is about to be
emitted. We separate the two readings by fitting one probe family to two targets at the same position:
the next candidate mention $M_t$ and the eventual final answer $F$. Because $M_t$ is a distractor by
construction in our main analysis, $M_t \neq F$, and a reader that tracks emission will score on $M_t$
but not on $F$. We fixed the analysis, its nulls, controls, and decision thresholds before fitting, and
recorded every amendment by commit in a time-stamped internal record; because that record is not public,
we call these analyses pre-specified rather than preregistered.

Our contributions are:
\begin{itemize}
  \item An emission-controlled probing protocol (ECDP) that fits $M_t$ and $F$ at the same position with
        target-specific nulls and a pre-specified emission gap $\Delta_{\mathrm{emit}}$.
  \item Evidence on DD in two model families, plus a Qwen3.5-4B result on a public reasoning task under its
        pre-specified fallback position set, that a probe fit before intermediate candidate mentions recovers
        the imminent mention far more accurately than the eventual answer (Section~\ref{sec:main}), and a pre-specified counterexample in which, under a different
        model, task, and reader, the final answer dominates (Section~\ref{sec:b3q}).
  \item An attempted matched cross-comparison, reported as incomplete together with its failure modes:
        on DD the imminent-mention probe did not train reliably under the full-prefix protocol in either model,
        which separates label validity from probe validity (Section~\ref{sec:mx}, Appendix~\ref{app:mx}).
  \item On DD, sensitivity controls showing that the reader recovers injected answer signals at the
        pre-mention positions, and a separate capacity study at the pre-rationale state finding increasing
        memorization without corresponding held-out gains (Section~\ref{sec:controls}); and a
        probe-specific natural-task non-detection whose sensitivity is checked only for an injected answer-slot
        code (Section~\ref{sec:boundary}).
  \item A transparent account of what causal interventions did and did not establish
        (Section~\ref{sec:causal}), and an explicit statement of the evidence tier of every result
        (Section~\ref{sec:limitations}).
\end{itemize}

Sections~\ref{sec:main} and~\ref{sec:controls} contain the main pre-specified results and the controls
that bound them. Section~\ref{sec:b3q} reports the counterexample, which limits how far the main result
generalizes, together with an attempted matched comparison that is incomplete and partly exploratory
(Section~\ref{sec:mx}). Sections~\ref{sec:boundary}, \ref{sec:e8}, and~\ref{sec:causal} are auxiliary: a natural-task
null, a reasoning-model setting whose verdict is held, and causal interventions that did not pass their own
checks. They are reported because they limit the interpretation, not because they add evidence for it.

We do not claim that hidden states carry no decision information. Prior work finds task-solving
information in CoT hidden states \citep{mehrafarin2026}, latent beliefs can be elicited from internal
states without supervision \citep{burns2023}, and our results are compatible with both. Our claim is
narrower: a strong answer-like readout can reflect imminent lexical emission, so decodability alone is
not evidence of a settled decision. Nor do we claim that imminent-emission dominance is a general property
of chain-of-thought readouts; Section~\ref{sec:b3q} reports a setting where it does not hold. Establishing that a decision state exists, or that it does not,
requires an intervention whose apparatus has been validated; ours was not (Section~\ref{sec:causal}).

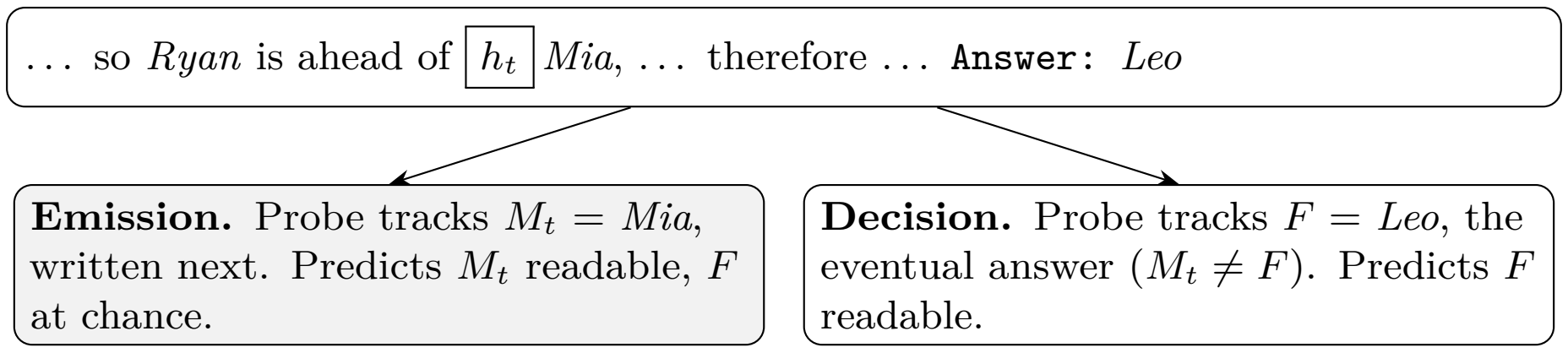
\begin{figure}[t]
\centering
\begin{tikzpicture}[every node/.style={font=\footnotesize}]
  \node[draw, rounded corners, inner sep=4pt, align=left, text width=0.72\textwidth] (r)
    {\ldots\ so \textit{Ryan} is ahead of \fbox{$h_t$}\,\textit{Mia}, \ldots\ therefore \ldots\ \texttt{Answer:} \textit{Leo}};
  \node[draw, rounded corners, fill=gray!10, align=left, text width=0.34\textwidth, anchor=north east] (m)
    at ([xshift=-1.5mm,yshift=-6mm]r.south)
    {\textbf{Emission.} Probe tracks $M_t$ = \textit{Mia}, written next. Predicts $M_t$ readable, $F$ at chance.};
  \node[draw, rounded corners, align=left, text width=0.34\textwidth, anchor=north west] (f)
    at ([xshift=1.5mm,yshift=-6mm]r.south)
    {\textbf{Decision.} Probe tracks $F$ = \textit{Leo}, the eventual answer ($M_t \neq F$). Predicts $F$ readable.};
  \draw[-{Stealth}] ([xshift=-1.2cm]r.south) -- (m.north);
  \draw[-{Stealth}] ([xshift=1.2cm]r.south) -- (f.north);
\end{tikzpicture}
\caption{Two readings of an answer probe at the token before an intermediate candidate mention. Standard
answer probing does not distinguish them. Names are illustrative.}
\label{fig:concept}
\end{figure}

\section{Related work}

The work closest to ours decodes the final answer from chain-of-thought states and asks \emph{when} it
becomes readable. ECDP asks \emph{what} a readout at a fixed position is reading, by fitting a second
target, the candidate about to be written, at the same position with the same reader. We summarise the
surrounding literature briefly.

\paragraph{Early answer decodability in chain of thought.} \citet{theater2026} disentangle model beliefs
from CoT, decoding final answers early on easy MMLU items and using this for early exit.
\citet{farahead2026} train a low-rank lens on CoT hidden states to predict subsequent tokens and final
answers, and report a myopic horizon in which precise answers become readable only one or two steps
before completion. \citet{commitment2026} estimate each step's causal importance from answer logits, find
a sharp commitment boundary that is decodable from activations, and exit early at it.
\citet{chainholds2026} report that answers can change under adversarial pressure while the reasoning
trace holds. \citet{cox2026} decode answers that are committed before the rationale and steer them
through activation directions; \citet{mehrafarin2026} find that when CoT fails, the solution can still
be recovered from hidden states. The closest probing studies among these primarily evaluate final-answer-related targets
at positions inside the rationale; we do not claim that every cited study uses the same reader, target,
or interpretation. We add a second target at the same position, the candidate about to be
named, and ask whether the answer readout survives it. The myopic horizon of \citet{farahead2026} and
the belief-versus-monitor gap of \citet{theater2026} are both statements about \emph{when} the final
answer becomes readable; the emission gap $\Delta_{\mathrm{emit}}$ is a statement about \emph{what} is
readable at a fixed position. Our steering attempt (Section~\ref{sec:causal}) did not pass its own
apparatus gate, so we do not use it as evidence against the causal findings of \citet{cox2026}.

\paragraph{Faithfulness of chain of thought.} A separate line of work asks whether the written
rationale reflects the computation that produced the answer. Explanations in few-shot prompting are
unreliable \citep{ye2022}; models produce plausible rationales that omit the feature that drove the
answer \citep{turpin2023}; the degree to which an answer depends on its rationale varies with model size
and task \citep{lanham2023}; and unfaithful reasoning occurs in the wild, not only under adversarial
prompts \citep{arcuschin2025}. These studies compare the rationale with the answer. We compare two
readouts of the same hidden state, and our result bears on faithfulness only indirectly: a probe that
tracks imminent emission reports what the rationale is about to say, which is not the same as what the
model will conclude.

\paragraph{Probe validity.} Linear probes were introduced as a way to read intermediate layers
\citep{alain2016}, and the question of what a successful probe shows has been studied since. Control
tasks \citep{hewitt2019} ask whether a probe reflects the representation or the probe's own capacity;
minimum-description-length probing measures how much the representation helps rather than whether a
probe can fit \citep{voita2020}; and surveys of probing classifiers list decodability, selectivity and
causal relevance as distinct properties that a probe result does not establish together
\citep{belinkov2022}. Amnesic probing removes a property from the representation and tests whether
behavior changes \citep{elazar2021,ravfogel2020}, behavioral evidence of the kind that
\citet{jacovi2020} argue is needed before an explanation can be called faithful. Causal abstraction gives a formal account of
when an intervention identifies a high-level variable \citep{geiger2021}. We follow this literature in
two ways: we use planted signals, label shuffles, position controls, and a capacity curve to separate
what the representation carries from what the reader can fit, and we keep a strict line between
decodability (Sections~\ref{sec:main} to~\ref{sec:boundary}) and causal relevance
(Section~\ref{sec:causal}).

\paragraph{Reading internal states.} Tuned lenses \citep{belrose2023} and Patchscopes
\citep{patchscopes2024} read hidden states by decoding them with the model itself; latent-knowledge and
self-knowledge studies find that internal states carry calibrated information about truth and about the
model's own answers \citep{burns2023,azaria2023,kadavath2022}. Our reader is deliberately smaller than
these tools so that its capacity can be controlled, and our question is not whether the state carries
answer information at all, but which of two competing targets it carries at the positions where answer
probes are usually reported. The wider CoT literature, from self-consistency and tree search
\citep{wang2023sc,yao2023tot} to bootstrapped and latent rationales \citep{zelikman2022star,zelikman2024quiet}
and step-level verification \citep{lightman2023}, treats intermediate text as computation; our result
is a caution about reading that computation off hidden states with a single-target probe.

\section{Setup}
\label{sec:setup}

\paragraph{Tasks.} \emph{DecisionDepth} (DD) is a synthetic multi-candidate reasoning task with three
families (rule chains, conditional code-name chains, and value intervals). Candidates are generated
symmetrically: each appears once in the state, in the same role, at a random slot, so no surface feature
identifies the answer. We use depth-2 items. As a public task we use BIG-Bench Hard Logical Deduction,
five- and seven-object subsets \citep{bbh2022}, converted so that candidates are object names. The
three-object subset is excluded because $K=3$ risks a ceiling.

\paragraph{Models and rollouts.} We use Qwen3.5-4B and Llama-3.1-8B-Instruct with thinking disabled and
a 2048-token rationale budget. The prompt explicitly asks the model to write a rationale and end with an \texttt{Answer:} line, so with
thinking disabled the rationale is ordinary requested output text, not a hidden reasoning trace.
On DD, Qwen3.5-4B is correct on \NQTroll{} of \NQTitems{} rollouts (chance \NQTchance). At the prompt
position, the probe recovers the final answer at \NQTprompt{}, near its chance baseline (\NQTchance). The
tested probe therefore detects no final-answer signal at that position; this does not establish that
generating the rationale is causally necessary. For
Llama-3.1-8B, direct answering is at \NLGdirect{} (chance \NLGchance) and the prompted rationale reaches
\NLGcot. On BBH Logical Deduction, Qwen3.5-4B scores \NBGdirect{} direct and \NBGcot{} with a rationale
(pilot set, chance \NBGchance). We analyze correct rollouts only, so $F$ equals the gold answer.

\paragraph{Representation and reader.} We take the layer-32 hidden state $h_t$, center it with
training-fold statistics, and apply parameter-free layer normalization. Candidates are embedded with the
same model at the same layer. The reader is a rank-4 bilinear scorer $s_k = (U x)^\top (V a_k)$ with
$2rd$ parameters for hidden size $d$ and rank $r=4$, with no projection: 20{,}480 for Qwen3.5-4B
($d=2560$) and 32{,}768 for Llama-3.1-8B ($d=4096$). It is trained with five item-level folds; weight decay is chosen on a calibration slice of
the training folds and the held-out fold is never used for selection. The two targets are fitted
separately, with the same reader family, folds, and selection rule.

\paragraph{Positions and targets.} For each rollout we locate every mention of a non-gold candidate in the
rationale and take the token immediately before it. The \emph{primary} set uses one such position per
item: the first distractor mention that occurs before the gold candidate is first named, so the final
answer has not yet appeared in the text. The \emph{secondary} set uses the first distractor mention
anywhere. At each position we fit two targets: $M_t$, the candidate named next, and $F$, the final answer.
All primary results are therefore conditional on correct rollouts that contain an eligible distractor
mention; rollouts that never name a distractor before the gold candidate do not enter the primary
analysis, and the results say nothing about them.

\paragraph{Nulls and the emission gap.} The two targets have different nulls. $F$ is uniform over $K$
candidates, so its null is $\mathbb{E}[1/K]$. $M_t$ is a non-gold candidate by construction, so a reader
that only learns ``not the gold one'' already gets $\mathbb{E}[1/(K-1)]$; we use this stricter null for
$M_t$. The pre-specified effect is
\begin{equation}
\Delta_{\mathrm{emit}} = \bigl[\mathrm{Acc}(M_t) - \mathbb{E}[1/(K-1)]\bigr]
 - \bigl[\mathrm{Acc}(F) - \mathbb{E}[1/K]\bigr],
\end{equation}
with a nominal 95\% within-seed interval from 10{,}000 item-level bootstrap resamples (the
pre-specified analysis used one reader seed; the interval does not cover training-seed variation). The verdict \textsc{Emission
Separated} requires $\mathrm{Acc}(M_t)$ at $z \geq 3$ and at least 0.15 above its null, $\mathrm{Acc}(F)$
below $z=3$ (a pre-specified screening threshold, not a conventional 5\% test or an equivalence
criterion), $\Delta_{\mathrm{emit}} \geq 0.10$, and a lower interval bound above zero, on at least
150 items. We compute $z=(\mathrm{Acc}-p_0)/\sqrt{p_0(1-p_0)/n}$, where $p_0$ is the
target-specific null and $n$ the number of held-out items; $z$ is a pre-specified screening statistic and
is not adjusted for multiplicity. Where the number of candidates $K_i$ varies across items, $p_0$ is the mean of the item-specific nulls; the item-specific binomial variance is then no larger than $p_0(1-p_0)/n$, so this pooled $z$ is conservative. With $K_i\ge 2$ an item-specific $z$ can exceed it by a factor of at most $\sqrt{2(1-p_0)}$, which leaves every below-threshold $F$ value reported here below $z=3$. Controls are a label shuffle, a random non-mention position in the same item, and a planted
positive control.

\section{Imminent mentions dominate the readout}
\label{sec:main}

\begin{figure}[t]
\centering
\begin{subfigure}{0.45\textwidth}\includegraphics[width=\linewidth]{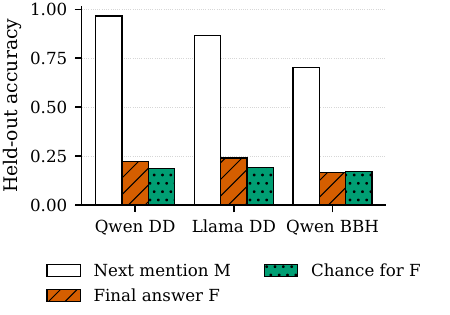}\caption{Accuracy by target}\end{subfigure}\hfill
\begin{subfigure}{0.53\textwidth}\includegraphics[width=\linewidth]{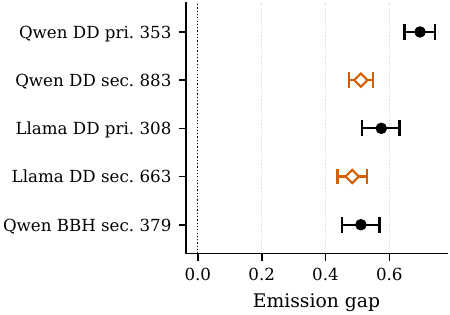}\caption{Emission gap with bootstrap interval}\end{subfigure}
\caption{Same positions and reader architecture, separately fitted to two targets. (a) Held-out accuracy for the next mention $M$ and the
final answer $F$ on each setting's pre-specified decision set; the third bar is the uniform null for $F$.
(b) Emission gap $\Delta_{\mathrm{emit}}$ with nominal 95\% within-seed item-bootstrap intervals from the median-gap seed
(training-seed variation is not included); filled markers are decision
sets, open markers the other set. The BBH primary set ($n=\NQBpriN$) is below the pre-specified 150-item
floor and is omitted.}
\label{fig:e1}
\end{figure}

\begin{table}[t]
\centering
\small
\resizebox{\textwidth}{!}{\begin{tabular}{lllrcccccl}
\toprule
Model & Task & Set & $n$ & Acc$_M$ (null) & Acc$_F$ (null) & $\Delta_{\mathrm{emit}}$ [95\% boot.] & Shuffled & Position & Verdict \\
\midrule
Qwen3.5-4B & DD & primary$^\dagger$ & 353 & .966 (.234) & .221 (.186) & .696 [.648, .744] & .144 & .204 & separated \\
Qwen3.5-4B & DD & secondary & 883 & .776 (.236) & .232 (.187) & .511 [.473, .549] & .181 & .199 & both readable \\
Llama-3.1-8B & DD & primary$^\dagger$ & 308 & .867 (.244) & .240 (.192) & .575 [.514, .632] & .204 & .179 & separated \\
Llama-3.1-8B & DD & secondary & 663 & .777 (.261) & .244 (.203) & .484 [.437, .531] & .187 & .208 & separated \\
Qwen3.5-4B & BBH-LD & primary & 61 & .377 (.219) & .131 (.178) & .157 [.009, .305] & .131 & .213 & underpowered \\
Qwen3.5-4B & BBH-LD & secondary$^\dagger$ & 379 & .705 (.211) & .166 (.173) & .511 [.451, .569] & .172 & .198 & separated \\
\bottomrule
\end{tabular}
}
\caption{Emission-controlled probing results. Nulls in parentheses: $\mathbb{E}[1/(K-1)]$ for $M$,
$\mathbb{E}[1/K]$ for $F$. $\dagger$ marks the decision set selected by the pre-specified rule for each setting (BBH uses the
secondary set because its primary set falls below 150 items). The pre-specified verdicts come from the
original single-seed analysis; accuracies and $z$ shown here are post-hoc medians over reader seeds 1--3,
and applying the same decision rule to them leaves every decision-set verdict unchanged. The bracketed
interval is a 10{,}000-resample item bootstrap for the median-gap seed; it is descriptive and does not
include training-seed variation. Shuffled: $M$ labels permuted, scored
against true $M$. Position: a random non-mention position in the same item (controls from seed 1).}
\label{tab:e1}
\end{table}

\begin{figure}[t]
\centering
\includegraphics[width=0.55\textwidth]{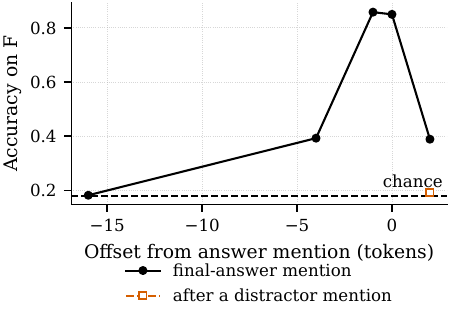}
\caption{Final-answer readout aligned to the first mention of the answer in the conclusion (Qwen3.5-4B,
DD, all correct rollouts). The readout peaks one token before the answer is written and decays after it.
The square is the readout two tokens after a distractor mention.}
\label{fig:align}
\end{figure}

\paragraph{High answer readouts sit just before the answer is written.} We first ask where the final
answer is readable at all. Aligning positions to the first mention of the answer in the conclusion
(Figure~\ref{fig:align}), the reader recovers $F$ at \NALmSixteen{} sixteen tokens before that mention,
\NALmFour{} four tokens before, and \NALmOne{} one token before; it falls back to \NALpTwo{} two tokens
after. Two tokens after a distractor mention it is at \NALdistrPTwo, at chance. At fixed fractions of the
rationale the answer stays near chance until the final step (\NQTprompt{} at the prompt, \NQTmid{} at the
midpoint, \NQTpre{} just before the answer line). A high answer readout therefore appears when the answer
is about to be written. This is the reading the rest of the section tests: if answer probes track
imminent emission, then just before a \emph{different} candidate is written they should track that
candidate instead.

\paragraph{Before a distractor mention, the readout follows the distractor.} Table~\ref{tab:e1} and
Figure~\ref{fig:e1} give the results. The pre-specified verdicts come from the original single-seed
analysis; the point estimates shown are post-hoc medians over three reader seeds, which leave those
verdicts unchanged, and intervals are from the seed with the median gap. On DD with Qwen3.5-4B, at the token before the
first distractor mention that precedes any mention of the answer, the reader recovers the next mention at
\NQDM{} against a null of \NQDchM{} ($z=\NQDzM$). From the same states it recovers the final answer at
\NQDF{} against \NQDchF{} ($z=\NQDzF$, below the $z=3$ threshold). The emission gap is \NQDD{}
[\NQDDlo, \NQDDhi]. Using every item's first distractor mention ($n=\NQDsecN$), which includes positions
after the answer has already been named, gives \NQDsecD{} [\NQDsecDlo, \NQDsecDhi]; here the final answer
is also weakly readable (\NQDsecF{} against \NQDsecchF, $z=\NQDseczF$, an excess of \NQDsecFexcess), so
this set is \textsc{Both Readable} by the pre-specified rule even though the gap stays large. In the
primary set the $F$ excess over its null is of similar size (\NQDFexcess) but below the $z=3$ threshold at
$n=\NQDn$ ($z=\NQDzF$), so we cannot say whether excluding post-answer positions removes the weak $F$
signal or only leaves it underpowered. Final-answer information can coexist with imminent-emission
information once the answer has appeared in the trajectory; the pre-specified pre-answer primary set
reduces this ambiguity and retains strong emission separation.

The same pattern holds for an independent model family. Llama-3.1-8B reads the next mention at \NLDM{}
and the final answer at \NLDF{}, a gap of \NLDD{} [\NLDDlo, \NLDDhi]. On BBH Logical Deduction,
the evidence is weaker in kind: the pre-answer primary set has only \NQBpriN{} items and falls below our 150-item
rule, so the pre-specified fallback is the secondary set, which may include positions after the answer has
been named. On that set Qwen3.5-4B reads the next mention at \NQBM{} and the final answer at \NQBF{},
below its uniform null, a gap of \NQBD{} [\NQBDlo, \NQBDhi]. The shuffle and position controls sit near their nulls in every
setting, so the $M_t$ readout is tied to the pre-mention position rather than to anything in the window.

\begin{figure}[t]
\centering
\includegraphics[width=0.55\textwidth]{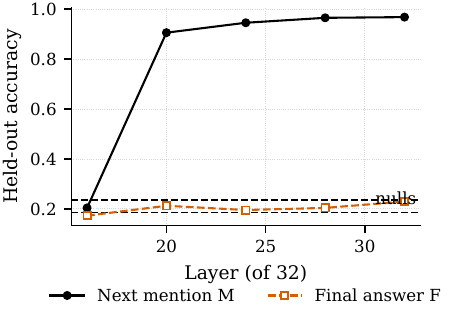}
\caption{Readout by layer at the primary positions (Qwen3.5-4B, DD; medians over three reader seeds).
Dashed lines are the nulls for $M$ (upper) and $F$ (lower). The next mention becomes readable from layer
20; the final answer stays below the $z=3$ threshold at every layer.}
\label{fig:layers}
\end{figure}

\paragraph{Not a last-layer effect.} Layer 32 is the last layer of Qwen3.5-4B, where the state must
predict the next token, so the separation could be specific to it. We refit the same reader at layers 16,
20, 24, and 28 on the same positions, using states from a separate teacher-forced extraction; at layer 32
this extraction reproduces the main result (gap \NLYgzero). At layer 16 neither target is readable
(\NLYSixteenM{} and \NLYSixteenF). From layer 20 on, the next mention is read at \NLYTwentyM{} (layer
20), \NLYTwentyFourM{} (24), and \NLYTwentyEightM{} (28) while the final answer stays at \NLYTwentyF,
\NLYTwentyFourF, and \NLYTwentyEightF; $F$ stays below $z=3$ at every layer (largest $z=\NLYmaxzF$), and
every layer where $M_t$ is readable is \textsc{Emission Separated} (Figure~\ref{fig:layers}). The layer
sweep was run on the primary positions only, so it says nothing about the layers at which the secondary
set's weak $F$ signal appears.

\section{On DD, the reader detects planted final-answer signals along the tested direction}
\label{sec:controls}

A low $\mathrm{Acc}(F)$ could mean the reader is too insensitive to a final-answer signal at these
positions. We test sensitivity with planted signals at the same positions; a separate capacity test on
the pre-rationale state asks whether a larger reader adds held-out accuracy. These controls rule out
blindness to the tested planted direction, not general insensitivity to a native final-answer code.

\begin{figure}[t]
\centering
\includegraphics[width=0.55\textwidth]{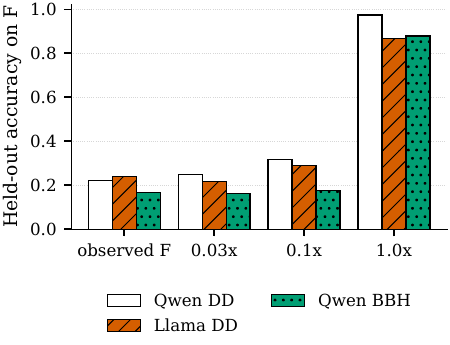}
\caption{Final-answer accuracy with no planting (observed) and with the gold embedding added at three
doses, on each setting's decision set (seed 1).}
\label{fig:planted}
\end{figure}

\paragraph{Planted signals.} We add the gold candidate's embedding to each state at doses of 1.0, 0.1,
and 0.03 and refit (Figure~\ref{fig:planted}; seed 1). On the Qwen DD decision set the reader recovers
$F$ at \NQDPoneX{} at full dose, \NQDPtenth{} at one tenth, and \NQDPthirtieth{} at one thirtieth. The
observed $F$ excess over its null, \NQDFexcess, is below the excess at the smallest dose
(\NQDPthirtiethexcess). Llama DD gives \NLDPoneX, \NLDPtenth, and \NLDPthirtieth{} against an observed
\NLDF. On BBH the full-dose plant is recovered (\NQBPoneX) but the one-tenth dose (\NQBPtenth) is at
its null (\NQBchF), so the sensitivity argument below is established on DD only. We read this
carefully. It shows the pipeline is not blind to answer information: on DD, a signal at one
tenth of the planted dose is already recovered above the observed level in both models. It does not show
that $F$ is absent: the data are consistent with an $F$ signal below about one thirtieth of the planted
dose in Qwen and between one thirtieth and one tenth in Llama, and a native answer code need not lie
along the candidate-embedding direction we plant. The comparison that matters is relative. The $M_t$
readout is comparable to planting the full answer signal, while $F$ is comparable to planting almost
none.

\begin{figure}[!htbp]
\centering
\begin{subfigure}{0.48\textwidth}\includegraphics[width=\linewidth]{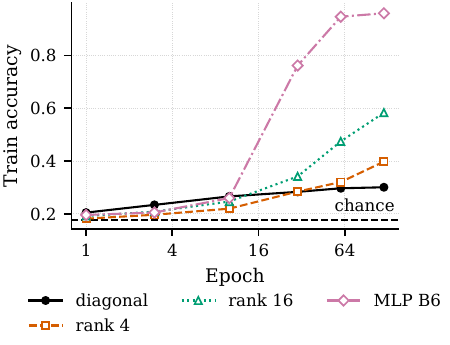}\caption{Train}\end{subfigure}\hfill
\begin{subfigure}{0.48\textwidth}\includegraphics[width=\linewidth]{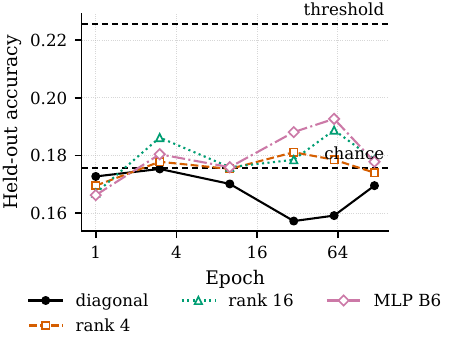}\caption{Held out}\end{subfigure}
\caption{Readers from a diagonal scorer to a \NECbsixM{}-parameter MLP trained to convergence without
early stopping (median of three seeds). Training accuracy grows with capacity, while held-out accuracy
stays between \NECmindev{} and \NECmaxdev{} and never reaches the pre-specified hidden-signal threshold in
any reader-by-checkpoint cell.}
\label{fig:capacity}
\end{figure}

\paragraph{Probe capacity.} We trained four readers of increasing capacity, from a diagonal scorer to a
\NECbsixM{}-parameter MLP, to read the final answer from the frozen layer-32 state representation before
any rationale (mean over the state's lines), for 120 epochs with no weight decay and no early stopping
(three seeds, median). Training accuracy rises with capacity from \NECtrainPzero{} to \NECtrainPthree{}
(Figure~\ref{fig:capacity}). Held-out accuracy stays between \NECmindev{} and \NECmaxdev{} in all
\NECcells{} reader-by-checkpoint cells, never reaching the pre-specified hidden-signal threshold of
\NECthr{} (null \NECchance). This test was deliberately lenient toward finding a hidden signal, since it
scans every reader and checkpoint without correction. At this before-rationale representation, added
capacity increases memorization without increasing held-out accuracy; the control does not test whether a
larger reader would recover $F$ at the pre-mention positions of Table~\ref{tab:e1}.

\FloatBarrier
\section{A natural-task non-detection with an injected-slot sensitivity control}
\label{sec:boundary}

A remaining objection is that our pipeline might fail to find early answer signals even where they exist
naturally. We asked whether the prompt-point state on MMLU \citep{mmlu2021} carries the eventual answer.
A first attempt used candidate-text embeddings, but MMLU options are semantically close, the planted
control failed its gate, and we recorded that attempt as uninformative. The one pre-specified redesign
changed only the candidate representation: four fixed slot codes, one per option position, scaled to the
same planted dose. This turns the reader into a low-rank answer-position probe while leaving the prompt,
layer, position, data split, and thresholds unchanged.

The sensitivity gate passed: the reader recovers a planted answer slot at \NESgone{} at full dose and
\NESptenth{} at one tenth. Opened only after that gate, the natural state yields \NESacc{} (chance
\NESchance, $z=\NESz$; balanced accuracy \NESbal; majority-slot baseline \NESmaj), so the pre-specified
verdict is \textsc{Final-Answer Signal Not Detected} (the internal pre-specification called this
label ``No Early Belief''; we rename it to avoid implying evidence of absence): at this position and layer, with this prompt, the
final-answer signal is not detected by this probe at the pre-specified threshold. The planted control
establishes sensitivity to that injected slot code, not the absence of a differently encoded native
signal, and the result is not an equivalence test. We do not take this as a contradiction of \citet{theater2026}, whose
positions, layers, prompts, and probes differ; it is a null control whose sensitivity was checked first.
We also do not compare its absolute accuracy with DD, since the candidate spaces differ.

\section{A reasoning model on lettered multiple choice (verdict held)}
\label{sec:e8}

To test the setting where early answer decodability has been reported, we followed the protocol of
\citet{theater2026} as closely as our resources allowed: DeepSeek-R1-Distill-Qwen-7B with thinking enabled
on MMLU-Redux, four lettered options, an attention-pooling reader, and the model's own final letter as the
label, keeping incorrect rollouts. Our reader pools a 64-token window rather than the full prefix, and we
use one 7B model rather than their larger ones. Of \NEEmmluN{} sampled items, \NEEmmluF{} produced a final
letter within 4{,}096 tokens (accuracy \NEEacc); on GPQA-Diamond only \NEEgpqaF{} of \NEEgpqaN{} did, too
few to analyze. Because one of this setting's pre-specified controls failed (below), none of its numbers
enters the main claim; the section is reported for completeness and as background for the attempted
matched comparison (Section~\ref{sec:mx}).

\paragraph{Early decodability was not reproduced.} A quarter of the way through the thinking trace the
reader recovers the final letter at \NEEqtr{} against a majority null of \NEEnullF{} ($z=\NEEqtrz$),
reaching \NEEend{} only at the end of the trace. By our pre-specified rule this setting does not reproduce
early belief, so the rest of this section is auxiliary evidence. We do not read it as a contradiction of
\citet{theater2026}; their full-prefix probe and larger models differ from ours, and a 64-token window is
a weaker reader by construction.

\paragraph{Before a non-final option, the readout follows that option (verdict held).} At the token before the first
mention of a non-final option that precedes any mention of the final one ($n=\NEEn$, layer 21 of 28), the
reader recovers the option about to be mentioned at \NEEM{} and the final letter at \NEEF{} ($z=\NEEzF$),
a gap of \NEED{} [\NEEDlo, \NEEDhi]; layer 14 gives \NEEMfourteen{} and \NEEFfourteen. As on DD, once the
final option has been mentioned the final letter becomes weakly readable (\NEEsecF, $z=\NEEseczF$). One
control did not behave as pre-specified: shuffling the next-option labels yields \NEEshuf, above the
uniform null of one third, because models tend to discuss options in order and the next-option labels are
not uniform. By our control rule this verdict is held. Using the shuffled accuracy as the null for the next
option, a post-hoc adjustment, the gap is still \NEEpost. We report this adjusted gap as a description of
the data, not as a replication: a setting whose shuffle control fails is not used as evidence that the
separation generalizes to reasoning models.

\section{A pre-specified counterexample: the final answer dominates on Qwen3-8B}
\label{sec:b3q}

The verdict of Section~\ref{sec:e8} was held, so we ran a follow-up whose rules were fixed
before its data were generated: the same dual-target comparison on a reasoning model, this time with the
attention-pooling reader of \citet{theater2026} over the full prefix. Two earlier attempts at this
follow-up are reported rather than dropped. On DeepSeek-R1-Distill-Qwen-7B the label-permutation null
failed its validity check, so that run has no statistics. On Llama-3.1-8B-Instruct the label extractor
failed its validity gate before any probe was fit: the model enumerates options from A to D and states
its conclusion without the letter, so the next letter mentioned follows enumeration order and does not
measure a choice. We keep that outcome as a limit on where letter-mention targets are valid. The third
attempt, Qwen3-8B, is therefore pre-specified for that model only, not as an independent test of the whole
study. It also shares a model family with R1-Distill-Qwen, so it is a replication across generation and
post-training rather than across architectures.

\paragraph{Design.} Probes are trained on rollouts of MMLU-Redux questions and evaluated on a confirmation
set of MMLU test questions that do not overlap MMLU-Redux, with thinking enabled and an empty system
prompt. $F$ is the model's own parsed answer, and correct and incorrect rollouts are both kept. $M_t$ is
the first option letter referenced in the next 64 tokens, found by fixed letter-reference patterns. The
decision positions are every thinking position at which $M_t$ is defined and differs from $F$. This is a
broader set than in Section~\ref{sec:main}, which uses only the token before a distractor's first
mention. The primary reader masks position~0 (the first chat-control token). We call this position-0
masking and do not treat it as a test of any attention-sink mechanism. Layers 22 and 31 were fixed by
relative depth. Each probe is fit with three seeds, and the baseline for each target is chosen on
validation data and then held fixed. For both targets it is the majority letter. The unit is the
question: per-position accuracies are averaged over seeds and then within questions, and intervals are
question-level bootstrap CIs at 97.5\% (Bonferroni over the two layers).

\paragraph{Result.} Every validity gate passed at both layers. Lift over the baseline is \NBQTtM{}
[\NBQTtMlo, \NBQTtMhi] for $M_t$ and \NBQTtF{} [\NBQTtFlo, \NBQTtFhi] for $F$ at layer 22, and \NBQToM{}
[\NBQToMlo, \NBQToMhi] and \NBQToF{} [\NBQToFlo, \NBQToFhi] at layer 31. The paired difference
$M_t - F$ is \NBQTtD{} [\NBQTtDlo, \NBQTtDhi] and \NBQToD{} [\NBQToDlo, \NBQToDhi]. Every seed at both
layers gives a difference between \NBQseedDmax{} and \NBQseedDmin. Both targets are readable, and the
final answer much more so. The unmasked control reader collapsed for some probes at layer 22 and is
reported separately. At layer 31 it trained normally and closely matched the masked reader, so the order
is not produced by masking.

\paragraph{Post-hoc analyses (not pre-specified).} The two targets have different label dynamics. $F$ is
constant within a rollout, while $M_t$ changes (mean within-question entropy \NBQwithinH{} bits). A
predictor that outputs one letter per question, chosen with oracle knowledge, would reach \NBQoracleM{} on
$M_t$, whereas a single constant suffices for $F$. The size of the lift difference therefore cannot be read
as a difference in stored information. Normalizing each lift by the headroom above its baseline leaves
the order unchanged (\NBQTthM{} for $M_t$, \NBQTthF{} for $F$ at layer 22). This is not a cross-entropy
comparison, which would need the probe logits, and we did not store them. Several observations suggest that
neither readout is trivial:
\begin{itemize}
  \item The $M_t$ probe tracks within-question changes. At positions where $M_t$ differs from its
        question's most frequent value it scores \NBQTtnonmodeM{}, against \NBQTtnonmodeMaj{} for the
        majority letter.
  \item The $F$ probe follows the model's answer rather than the gold answer. On the \NBQnWrong{}
        questions answered incorrectly, it outputs the model's answer at \NBQTtwrongF{} and the gold
        letter at \NBQTtwrongGold{}.
  \item The $F$ readout is present before $F$ is written. Its lift is \NBQTtpreF{}
        [\NBQTtpreFlo, \NBQTtpreFhi] at positions before any option letter has been referenced. Within
        the \NBQTtwN{} questions that have positions on both sides of $F$'s first mention, $F$ accuracy
        rises from \NBQTtbefF{} before that mention to \NBQTtaftF{} after it.
  \item Copying the most recently mentioned letter gives a lift of \NBQruleF{} for $F$ and \NBQruleM{}
        for $M_t$, so the $F$ readout is not a copy of recent text.
\end{itemize}
The $M_t$ probe also outputs the final letter at \NBQTtMreadsF{} of $M_t \neq F$ positions. In this
setting the final answer is a strong competitor to the imminent mention, the reverse of
Section~\ref{sec:main}.

\paragraph{What this changes.} The dominance of the imminent mention on DD and BBH is a property of those
settings, not a general law. Model, task, decision positions, and reader all differ between them and this
section, so we cannot say which factor reverses the order. Section~\ref{sec:mx} reports an attempt to
separate them.

\subsection{An attempted matched cross-comparison (incomplete)}
\label{sec:mx}

We attempted to separate model from task by crossing two models (Qwen3.5-4B and Qwen3-8B) with two tasks
(DD and MMLU) under the protocol of this section: thinking enabled, every thinking position at which
$M_t$ is defined and differs from $F$, the position-0-masked full-prefix reader, layers fixed by relative
depth (19 and 27 for Qwen3.5-4B), and the same gates and statistics. The Qwen3-8B MMLU cell is the result
above; it was already known when the cross was designed, so it is not a new test. The design, gates, and the outcome of every cell are in Appendix~\ref{app:mx}. The comparison is
incomplete, and we report what it does and does not support.

\paragraph{DD cells: no valid statistics.} For Qwen3.5-4B on DD, a check that each imminent-mention probe
trains, added after the smoke runs failed and fixed before it ran, failed in a full-sample diagnostic: training loss improved for \NMXcOneTrainOk{} of
\NMXcOneHeads{} imminent-mention probes and validation loss for \NMXcOneValOk. We closed this cell as an
incomplete evaluation without further tuning. For Qwen3-8B on DD, the labels passed their own validity
checks, but at both layers one imminent-mention seed ended with a higher training loss than it started
(\NMXcThreeTtsZero{} to \NMXcThreeTtsFin{} at layer 22, \NMXcThreeTosZero{} to \NMXcThreeTosFin{} at
layer 31), so by the pre-specified rule the cell has no statistics. Across that cell's masked imminent-mention
probes the final training loss stayed between \NMXcThreeMlo{} and \NMXcThreeMhi{}, at or slightly above the
uniform-guess cross-entropy measured on the Qwen3.5-4B DD items (\NMXddCE{}), a report-only observation, while final-answer probes reached
\NMXcThreeFlo{} to \NMXcThreeFhi. We did not relax the gate for this cell or compute its accuracies.
These failures show that this reader and training procedure did not reliably extract the DD imminent
mention; they are not evidence that the hidden states lack it. Their cause is untested: candidates include
the DD labels in combination with the reader, and the DD-specific label path (slot-index labels with a
logit mask over $K$ candidates and $K$-stratified baselines), which the MMLU cells never exercise and
which was checked only on synthetic inputs. A cell can pass its label checks and still fail as a probe.

\paragraph{Qwen3.5-4B on MMLU: exploratory.} The training checks that gated this cell's main run also
failed: the smoke runs, and the full-sample diagnostic added after them (fixed before it ran), in which
\NMXcTwoTrainOk{} of \NMXcTwoHeads{} imminent-mention probes improved training loss and \NMXcTwoValOk{}
improved validation loss; the one failure, an unmasked control-reader probe at layer 27, diverged late in
training. That failure is kept on record. After seeing it we wrote a separate protocol, fixed
before the C2x or C3 confirmation results were examined (the C4 results were already known), that runs the main probe unchanged and reports the
original training gate without using it as a gate; all other gates and statistics are unchanged. The protocol applies only to this cell; it was sealed
before C3 was opened, and C3 was scored under the original rules. Results
under this protocol are post-hoc and exploratory. Both targets are decodable, and the final answer more
strongly: the difference $M_t - F$ is \NMXNiD{} [\NMXNiDlo, \NMXNiDhi] at layer 19 and \NMXTsD{}
[\NMXTsDlo, \NMXTsDhi] at layer 27 over \NMXn{} questions. Of \NMXtotal{} target probes, \NMXvalid{} met the
validity criteria (gradient flow, validation improvement, no attention collapse). The three invalid probes are all at layer 27: two belong to the unmasked control reader, which collapsed
at that layer, so the masking control is not available there, and one is the masked final-answer probe of
one seed, which did not improve its validation loss; and that seed's difference is
\NMXTsseedTwoD{} against \NMXTsseedOtherDa{} and \NMXTsseedOtherDb{} for the other two; the layer-27
average should not be read on its own. As fixed in the protocol, no probe was excluded and no sensitivity
analysis without it was run.

\paragraph{Model contrast on MMLU.} In an exploratory paired contrast on the \NMXpairN{} confirmation
questions shared by the two MMLU cells (of \NMXn{} in C2x and \NBQn{} in C4),
the paired difference in $M_t - F$ between Qwen3-8B (layer 22) and Qwen3.5-4B (layer 19) is \NMXconE{}
[\NMXconElo, \NMXconEhi]. This interval does not detect a difference between the models; it does not show
that they are equivalent. The deeper-layer contrast, \NMXconL{} [\NMXconLlo, \NMXconLhi], depends on the
invalid probe described above, and we do not use it as evidence of a model difference. A model here means
weights, architecture, chat template, inference engine, and activation-extraction path together.

\paragraph{What remains.} Because neither DD cell has statistics, the task contrast within either model
and the model-by-task interaction cannot be estimated. The main-section result (imminent mention dominant
on DD for Qwen3.5-4B) and the MMLU results here differ in task, positions, reader, thinking mode, label definition, null, and prompt at once,
so they do not identify which change reverses the order.

\section{Causal interventions}
\label{sec:causal}

Decodability is a property of a reader; whether the decoded variable controls behavior is a separate,
causal question \citep{belinkov2022,geiger2021}. We attempted to answer it and report what happened.

\paragraph{Gradient steering (uninterpretable).} We steered layer-\NEFlayer{} states along the exact
input-gradient of the probe's preference for a target candidate, at a magnitude fixed on a calibration
split ($\alpha=\NEFalpha$ of the state norm). Replay without intervention reproduced the original next
mention at \NEFreplay. The pre-specified apparatus check, steering at the answer slot toward a different
candidate, changed the answer in none of \NEFn{} items ($\Delta=\NEFgtwo$). The hook did act: at
distractor positions it changed the next token in \NEFauditSteer{} of \NEFn{} items. Because the apparatus
gate failed, the pre-specified verdict is \textsc{Uninterpretable}, and we draw no causal conclusion in
either direction.

\paragraph{Full-state interchange (exploratory).} After seeing that result we ran one exploratory
intervention, pre-specified as such before any code was written, with a fixed rule that the causal branch
would close regardless of outcome. Within the same trajectory we replaced the layer-20 state at the
position before a mention of $B$ with the state from the position before a mention of another candidate
$C'$ (Table~\ref{tab:e4b}). The next mention became $C'$ in \NEBswap{} of items, against \NEBrand{} when
the donor was a random position in the same trajectory: a difference of \NEBdf{} [\NEBdflo, \NEBdfhi]. The
reverse direction gives \NEBdr{} [\NEBdrlo, \NEBdrhi]. Using a state from another mention of the same
candidate $B$ left $B$ in place (\NEBsameKeepB), and a perturbation matched in norm did nothing
(\NEBnorm), so the effect follows candidate identity rather than position or perturbation size. Across
these large changes in the imminent mention, the eventual answer changed in \NEBfinSwap{} of swapped items
versus \NEBfinNone{} without intervention.

We report three things from this and no more. Full-state interchange controls the imminent mention
strongly. It cannot be compared with gradient steering, whose apparatus failed; if a difference between
the two were ever established, distributed or nonlinear codes, off-manifold steering, or a rank-4 probe
that misses part of the causal subspace would each be an untested explanation. And large changes in the imminent mention were accompanied by little observed change in the eventual
answer. This last observation is exploratory and not an equivalence test: no effect bound was fixed in
advance, so it cannot be read as evidence that the final answer is insensitive to the swapped state. We
also note the strongest objection: the full state just before a mention necessarily carries what
downstream layers need to produce the next token, so swap success on its own is close to expected.
Identifying an isolated decision variable would require the stronger conditions of a causal abstraction
\citep{geiger2021}, which we do not claim to meet.

\begin{table}[t]
\centering
\small
\begin{tabular}{lrcc}
\toprule
Donor state & $n$ & $P(C')$ & $P(B)$ \\
\midrule
none & 397 & .000 & .998 \\
identity & 397 & .000 & .998 \\
swap ($C'$ state) & 397 & .700 & .277 \\
random position & 397 & .086 & .524 \\
same cand., other mention & 162 & .006 & .982 \\
norm-matched noise & 397 & .003 & .990 \\
\bottomrule
\end{tabular}

\caption{Exploratory full-state interchange at layer 20 (Qwen3.5-4B, DD). The state at the position
before a mention of $B$ is replaced by a donor state from the same trajectory; columns give the
probability that the next mention is $C'$ or $B$. The same-candidate control is defined only for items
with a second mention of $B$.}
\label{tab:e4b}
\end{table}

\section{Discussion}

\paragraph{What answer probing measures.} A probe's accuracy is a property of a target at a position, not
of ``the answer'' in general. At positions immediately before candidate mentions, as studied here, the
next emitted candidate is a strong competing target on DD and BBH, while in the MMLU setting of Section~\ref{sec:b3q} (Qwen3-8B; Qwen3.5-4B exploratory), which also
differs in reader, positions, and labels, the final answer dominates. Evaluations at such positions that read a high
answer-probe score as a formed decision should report the imminent-mention target at the same position and show that the answer target
survives it.

\paragraph{What this does not show.} Two readings of our result are stronger than the evidence. The
first is that the model has no final-answer information before it writes the answer. The planted control
bounds a native signal only relative to a chosen injection direction and dose; the capacity study finds no
held-out improvement among the tested readers at the pre-rationale state, and it does not rule out
final-answer information encoded in another form or recoverable by a different reader, particularly at
the pre-mention positions; and the secondary DD
set shows that a weak $F$ readout does appear once the answer has been named. In the Qwen3-8B setting a
strong $F$ readout is present before any option is written. The second is that the
imminent mention is causally upstream of the decision. Our steering apparatus failed its gate, and the
interchange was exploratory. The defensible statement is the one in the title: a readout that recovers
an answer candidate is not, by itself, evidence that a decision has been made. At the selected DD and BBH
positions, separately fitted readers recover the imminent mention much more accurately than the eventual
answer. In the Qwen3-8B setting the reverse holds, and a strong final-answer readout there is still not
evidence of when the model committed. Whether the competing signal explains successful final-answer probes
elsewhere remains untested.

\paragraph{Relation to early-belief results.} The early decodability reported by \citet{theater2026}
and the commitment boundary of \citet{commitment2026} are measured on lettered multiple-choice items
with full-prefix or logit-based readers, in settings where the answer letter is rarely written before
the conclusion. Our positions are chosen so that a non-answer candidate is about to be written. The two
sets of findings can both hold: an answer may be readable early on easy items, and at the same time a
probe placed before a candidate mention may be reading the mention. What our result changes is the
inference, not the measurement. Section~\ref{sec:b3q} moves partway toward a matched comparison, using
their full-prefix attention reader on MMLU, and finds the final answer dominant there for Qwen3-8B (pre-specified) and, in an exploratory post-hoc
evaluation at layer 19 only, for Qwen3.5-4B. It is
not a matched comparison, because task changed together with the reader and the positions. The cross design
that would separate these factors was attempted and is incomplete (Section~\ref{sec:mx}).

\paragraph{A probe that does not train is not a null.} Probe decodability depends on the validity of the
readout procedure. On DD under the full-prefix protocol, our imminent-mention probes did not train
reliably in either model (Section~\ref{sec:mx}, Appendix~\ref{app:mx}). We treat these failures as ones of the reader, its
optimization, or their interaction with the labels, and failed probe optimization should not be
interpreted as evidence that the corresponding information is absent from the model's hidden states.
Reporting it as absence of information would be the inverse of the error this paper is about: an untrained
probe is no more evidence that a decision is absent than a trained one is evidence that it is present.

\paragraph{A practical protocol.} The controls we found necessary are cheap: fit the next-mention target
alongside the answer target, use target-specific nulls, include a position control and a label shuffle,
pre-specify the thresholds, and calibrate sensitivity with planted signals before interpreting a null.
Report failed apparatus checks rather than dropping them; a probe that clears a statistical gate is not
thereby a faithful account of the computation \citep{hewitt2019,belinkov2022,voita2020}.

\section{Limitations and evidence tiers}
\label{sec:limitations}

\paragraph{Evidence tiers.} We distinguish five tiers. \emph{Primary pre-specified}: the decision
sets in Table~\ref{tab:e1} (Qwen DD primary, Llama DD primary, BBH secondary), each passing every
pre-specified criterion. The primary analysis definition, not the reported numbers, is pre-specified:
Table~\ref{tab:e1} reports subsequently added three-seed robustness summaries rather than the original
pre-specified numerical estimates. The decision rule and the original single-seed analysis were fixed in a
non-public internal record; the three-seed medians and median-gap-seed intervals reported here are
robustness summaries added afterward; the auxiliary layer, planted-signal, and capacity controls have the narrower,
setting-specific scopes stated in Sections~\ref{sec:main}--\ref{sec:controls}. \emph{Limited}: the Qwen
DD secondary set, which is \textsc{Both Readable} by seed median; the BBH primary set, which has
\NQBpriN{} items against a floor of 150; and the MMLU prompt-point null, which is a null at one position
and layer under one prompt. \emph{Exploratory, post-hoc}: the Qwen3.5-4B
MMLU cell of the matched comparison (Section~\ref{sec:mx}), run under a protocol written after the training checks that gated its main run failed; that failure stands. \emph{Not supporting}: the two DD cells of the
matched comparison, which have no statistics; the reasoning-model setting
(Section~\ref{sec:e8}), whose shuffle control failed; the gradient-steering attempt, which failed its
apparatus gate; and the full-state interchange, which was exploratory and is not an equivalence test.
\emph{Pre-specified for one model}: the Qwen3-8B counterexample (Section~\ref{sec:b3q}). Its rules were
fixed before its data existed, but after two failed attempts at the same follow-up, so it is not a
study-level independent test. Its confound analyses are post-hoc and report-only. The abstract's
emission-gap claims come from the primary tier, its counterexample from this tier, and its Qwen3.5-4B MMLU
statement from the exploratory tier; its planted-signal and capacity
statements are auxiliary controls with the scopes above; its intervention statements are limitations.

\paragraph{Scope.} Our main analysis conditions on correct rollouts, where $F$ equals the gold answer;
incorrect DD rollouts were too few to analyze, so we cannot say whether the separation holds there when the
model's eventual answer is wrong. The Qwen3-8B setting keeps incorrect rollouts (\NBQnWrong{} questions), but
it uses a different protocol. We study four models of 4B to 8B parameters, one synthetic task, and public multiple-choice tasks; three of
the models share the Qwen lineage, and the DD layer analysis covers Qwen3.5-4B only. DD candidates are symmetric by construction; on natural tasks
the order in which a model discusses candidates is not uniform, as the failed shuffle control in
Section~\ref{sec:e8} suggests, and order-aware nulls would be needed there. The pre-specified analysis
used a single reader seed; the three-seed medians reported here left every decision-set verdict unchanged
but turned the Qwen DD secondary set from \textsc{Emission Separated} to \textsc{Both Readable}. The
Llama DD secondary set, with an $F$ excess of similar size, stays \textsc{Emission Separated} by the same
rule; the verdict is sensitive to the significance threshold there. The planted control
bounds the $F$ signal only relative to planted doses along one direction.

\paragraph{Controls not yet run.} Three experiments would strengthen or overturn the interpretation and
have not been performed. (i) A complete matched cross-comparison. Our attempt (Section~\ref{sec:mx}) has no valid DD cell, so
the reversal cannot be attributed to model or to task and positions. The held-out cross-entropy reduction
per target, pre-specified as a report-only item, was not computed for the cells reported here. Completing it would first require a
reader and training procedure that demonstrably learns the DD imminent-mention target, fixed before the
cross is rerun; reader type would be a separate axis, varied alone. A comparison with matched label
dynamics, such as held-out cross-entropy against each target's own baseline, belongs with it. (ii) Incorrect rollouts: a difficulty level fixed in advance at which enough
rollouts are wrong to fit both targets, reported conditional on correctness and pooled. (iii) A causal
intervention with a passing on-manifold positive control, with the change rates of the imminent mention
and of the final answer reported as a paired interval against an effect bound fixed before the run. Until
(iii) exists, nothing in this paper bears on whether the imminent mention causes the decision. Three
further controls are smaller in scope but also not run: (iv) order-aware nulls for natural tasks, where
the next-option labels are not uniform; (v) more model families and sizes, and the layer sweep on Llama
and BBH; (vi) a full-prefix reader on the reasoning-model setting of Section~\ref{sec:e8}, so that the
64-token window used there can be compared with the full-prefix reader used in Section~\ref{sec:b3q}.

\section{Conclusion}

Near intermediate candidate mentions on DecisionDepth, a probe reads the candidate about to be written far
more accurately than the eventual answer in Qwen3.5-4B and Llama-3.1-8B, for a reader that detects planted
answer signals at one tenth of full dose on that task; on BIG-Bench Hard the same ordering appears for
Qwen3.5-4B on the broader secondary position set, while the pre-answer primary set is underpowered. In a pre-specified Qwen3-8B
evaluation on MMLU with a full-prefix reader over every thinking position, the order reverses: both targets
are readable, and the final answer much more so. Neither pattern is a general law of chain-of-thought
readouts. Under the MMLU protocol, an exploratory, post-hoc evaluation of Qwen3.5-4B also reads the final answer
more strongly at layer 19; its layer-27 average is not evidence on its own because it includes one invalid
final-answer probe. Which target dominates depends on the setting; our attempted matched comparison is
incomplete, because the imminent-mention probe did not train reliably on DD under that protocol, so we do not
identify which factor reverses the order. A high or low answer readout is not, on
its own, evidence of when a decision is made.

\bibliographystyle{plainnat}
\bibliography{refs}

\begin{thebibliography}{32}
\providecommand{\natexlab}[1]{#1}
\providecommand{\url}[1]{\texttt{#1}}
\expandafter\ifx\csname urlstyle\endcsname\relax
  \providecommand{\doi}[1]{doi: #1}\else
  \providecommand{\doi}{doi: \begingroup \urlstyle{rm}\Url}\fi

\bibitem[Alain and Bengio(2016)]{alain2016}
Guillaume Alain and Yoshua Bengio.
\newblock Understanding intermediate layers using linear classifier probes,
  2016.
\newblock arXiv:1610.01644.

\bibitem[Arcuschin et~al.(2025)Arcuschin, Janiak, Krzyzanowski, Rajamanoharan,
  Nanda, and Conmy]{arcuschin2025}
Iv{\'a}n Arcuschin, Jett Janiak, Robert Krzyzanowski, Senthooran Rajamanoharan,
  Neel Nanda, and Arthur Conmy.
\newblock Chain-of-thought reasoning in the wild is not always faithful, 2025.
\newblock arXiv:2503.08679.

\bibitem[Azaria and Mitchell(2023)]{azaria2023}
Amos Azaria and Tom Mitchell.
\newblock The internal state of an {LLM} knows when it's lying, 2023.
\newblock arXiv:2304.13734; Findings of EMNLP 2023.

\bibitem[Belinkov(2022)]{belinkov2022}
Yonatan Belinkov.
\newblock Probing classifiers: Promises, shortcomings, and advances.
\newblock \emph{Computational Linguistics}, 48\penalty0 (1):\penalty0 207--219,
  2022.
\newblock arXiv:2102.12452.

\bibitem[Belrose et~al.(2023)Belrose, Furman, Smith, Halawi, Ostrovsky,
  McKinney, Biderman, and Steinhardt]{belrose2023}
Nora Belrose, Zach Furman, Logan Smith, Danny Halawi, Igor Ostrovsky, Lev
  McKinney, Stella Biderman, and Jacob Steinhardt.
\newblock Eliciting latent predictions from transformers with the tuned lens,
  2023.
\newblock arXiv:2303.08112.

\bibitem[Boppana et~al.(2026)Boppana, Ma, Loeffler, Sarfati, Bigelow, Geiger,
  Lewis, and Merullo]{theater2026}
Siddharth Boppana, Annabel Ma, Max Loeffler, Raphael Sarfati, Eric Bigelow,
  Atticus Geiger, Owen Lewis, and Jack Merullo.
\newblock Reasoning theater: Disentangling model beliefs from chain-of-thought.
\newblock In \emph{Proceedings of the 43rd International Conference on Machine
  Learning}, volume 306 of \emph{Proceedings of Machine Learning Research},
  pages 9152--9177, 2026.
\newblock arXiv:2603.05488.

\bibitem[Burns et~al.(2023)Burns, Ye, Klein, and Steinhardt]{burns2023}
Collin Burns, Haotian Ye, Dan Klein, and Jacob Steinhardt.
\newblock Discovering latent knowledge in language models without supervision.
\newblock In \emph{International Conference on Learning Representations}, 2023.
\newblock arXiv:2212.03827.

\bibitem[Cox et~al.(2026)Cox, Kianersi, and Garriga-Alonso]{cox2026}
Kyle Cox, Darius Kianersi, and Adri{\`a} Garriga-Alonso.
\newblock Post-hoc reasoning in chain of thought: Decoding and steering
  pre-committed answers, 2026.
\newblock arXiv:2603.01437.

\bibitem[Elazar et~al.(2021)Elazar, Ravfogel, Jacovi, and Goldberg]{elazar2021}
Yanai Elazar, Shauli Ravfogel, Alon Jacovi, and Yoav Goldberg.
\newblock Amnesic probing: Behavioral explanation with amnesic counterfactuals.
\newblock \emph{Transactions of the Association for Computational Linguistics},
  9:\penalty0 160--175, 2021.
\newblock arXiv:2006.00995.

\bibitem[Geiger et~al.(2021)Geiger, Lu, Icard, and Potts]{geiger2021}
Atticus Geiger, Hanson Lu, Thomas Icard, and Christopher Potts.
\newblock Causal abstractions of neural networks, 2021.
\newblock arXiv:2106.02997; NeurIPS 2021.

\bibitem[Ghandeharioun et~al.(2024)Ghandeharioun, Caciularu, Pearce, Dixon, and
  Geva]{patchscopes2024}
Asma Ghandeharioun, Avi Caciularu, Adam Pearce, Lucas Dixon, and Mor Geva.
\newblock Patchscopes: A unifying framework for inspecting hidden
  representations of language models.
\newblock In \emph{Proceedings of the 41st International Conference on Machine
  Learning}, volume 235 of \emph{Proceedings of Machine Learning Research},
  2024.
\newblock arXiv:2401.06102.

\bibitem[Hendrycks et~al.(2021)Hendrycks, Burns, Basart, Zou, Mazeika, Song,
  and Steinhardt]{mmlu2021}
Dan Hendrycks, Collin Burns, Steven Basart, Andy Zou, Mantas Mazeika, Dawn
  Song, and Jacob Steinhardt.
\newblock Measuring massive multitask language understanding.
\newblock In \emph{International Conference on Learning Representations}, 2021.
\newblock arXiv:2009.03300.

\bibitem[Hewitt and Liang(2019)]{hewitt2019}
John Hewitt and Percy Liang.
\newblock Designing and interpreting probes with control tasks.
\newblock In \emph{Proceedings of EMNLP-IJCNLP}, pages 2733--2743, 2019.
\newblock arXiv:1909.03368.

\bibitem[Jacovi and Goldberg(2020)]{jacovi2020}
Alon Jacovi and Yoav Goldberg.
\newblock Towards faithfully interpretable {NLP} systems: How should we define
  and evaluate faithfulness?
\newblock In \emph{Proceedings of ACL}, pages 4198--4205, 2020.
\newblock arXiv:2004.03685.

\bibitem[Kadavath et~al.(2022)Kadavath, Conerly, Askell, Henighan, Drain,
  Perez, et~al.]{kadavath2022}
Saurav Kadavath, Tom Conerly, Amanda Askell, Tom Henighan, Dawn Drain, Ethan
  Perez, et~al.
\newblock Language models (mostly) know what they know, 2022.
\newblock arXiv:2207.05221.

\bibitem[Kojima et~al.(2022)Kojima, Gu, Reid, Matsuo, and Iwasawa]{kojima2022}
Takeshi Kojima, Shixiang~Shane Gu, Machel Reid, Yutaka Matsuo, and Yusuke
  Iwasawa.
\newblock Large language models are zero-shot reasoners.
\newblock In \emph{Advances in Neural Information Processing Systems}, 2022.
\newblock arXiv:2205.11916.

\bibitem[Lanham et~al.(2023)Lanham, Chen, Radhakrishnan, Steiner, Denison,
  et~al.]{lanham2023}
Tamera Lanham, Anna Chen, Ansh Radhakrishnan, Benoit Steiner, Carson Denison,
  et~al.
\newblock Measuring faithfulness in chain-of-thought reasoning, 2023.
\newblock arXiv:2307.13702.

\bibitem[Li et~al.(2026)Li, Krishnan, and Padman]{chainholds2026}
Yubo Li, Ramayya Krishnan, and Rema Padman.
\newblock The chain holds, the answer folds: Trace-answer dissociation in
  reasoning models under adversarial pressure, 2026.
\newblock arXiv:2605.29087.

\bibitem[Lightman et~al.(2023)Lightman, Kosaraju, Burda, Edwards, Baker, Lee,
  Leike, Schulman, Sutskever, and Cobbe]{lightman2023}
Hunter Lightman, Vineet Kosaraju, Yura Burda, Harri Edwards, Bowen Baker, Teddy
  Lee, Jan Leike, John Schulman, Ilya Sutskever, and Karl Cobbe.
\newblock Let's verify step by step, 2023.
\newblock arXiv:2305.20050; ICLR 2024.

\bibitem[Mehrafarin et~al.(2026)Mehrafarin, Parekh, and
  Konstas]{mehrafarin2026}
Houman Mehrafarin, Amit Parekh, and Ioannis Konstas.
\newblock When chain-of-thought fails, the solution hides in the hidden states,
  2026.
\newblock arXiv:2604.23351.

\bibitem[Ravfogel et~al.(2020)Ravfogel, Elazar, Gonen, Twiton, and
  Goldberg]{ravfogel2020}
Shauli Ravfogel, Yanai Elazar, Hila Gonen, Michael Twiton, and Yoav Goldberg.
\newblock Null it out: Guarding protected attributes by iterative nullspace
  projection.
\newblock In \emph{Proceedings of ACL}, pages 7237--7256, 2020.
\newblock arXiv:2004.07667.

\bibitem[Scalena et~al.(2026)Scalena, Candussio, Bortolussi, Fersini, Nissim,
  and Sarti]{commitment2026}
Daniel Scalena, Sara Candussio, Luca Bortolussi, Elisabetta Fersini, Malvina
  Nissim, and Gabriele Sarti.
\newblock Beyond the commitment boundary: Probing epiphenomenal
  chain-of-thought in large reasoning models, 2026.
\newblock arXiv:2606.13603.

\bibitem[Suzgun et~al.(2022)Suzgun, Scales, Sch{\"a}rli, Gehrmann, Tay, Chung,
  Chowdhery, Le, Chi, Zhou, and Wei]{bbh2022}
Mirac Suzgun, Nathan Scales, Nathanael Sch{\"a}rli, Sebastian Gehrmann, Yi~Tay,
  Hyung~Won Chung, Aakanksha Chowdhery, Quoc~V. Le, Ed~H. Chi, Denny Zhou, and
  Jason Wei.
\newblock Challenging {BIG-Bench} tasks and whether chain-of-thought can solve
  them, 2022.
\newblock arXiv:2210.09261.

\bibitem[Turpin et~al.(2023)Turpin, Michael, Perez, and Bowman]{turpin2023}
Miles Turpin, Julian Michael, Ethan Perez, and Samuel~R. Bowman.
\newblock Language models don't always say what they think: Unfaithful
  explanations in chain-of-thought prompting.
\newblock In \emph{Advances in Neural Information Processing Systems}, 2023.
\newblock arXiv:2305.04388.

\bibitem[Voita and Titov(2020)]{voita2020}
Elena Voita and Ivan Titov.
\newblock Information-theoretic probing with minimum description length.
\newblock In \emph{Proceedings of EMNLP}, pages 183--196, 2020.
\newblock arXiv:2003.12298.

\bibitem[Wang et~al.(2023)Wang, Wei, Schuurmans, Le, Chi, Narang, Chowdhery,
  and Zhou]{wang2023sc}
Xuezhi Wang, Jason Wei, Dale Schuurmans, Quoc~V. Le, Ed~H. Chi, Sharan Narang,
  Aakanksha Chowdhery, and Denny Zhou.
\newblock Self-consistency improves chain of thought reasoning in language
  models.
\newblock In \emph{International Conference on Learning Representations}, 2023.
\newblock arXiv:2203.11171.

\bibitem[Wei et~al.(2022)Wei, Wang, Schuurmans, Bosma, Ichter, Xia, Chi, Le,
  and Zhou]{wei2022}
Jason Wei, Xuezhi Wang, Dale Schuurmans, Maarten Bosma, Brian Ichter, Fei Xia,
  Ed~H. Chi, Quoc~V. Le, and Denny Zhou.
\newblock Chain-of-thought prompting elicits reasoning in large language
  models.
\newblock In \emph{Advances in Neural Information Processing Systems}, 2022.
\newblock arXiv:2201.11903.

\bibitem[Xu et~al.(2026)Xu, Yu, Meng, and Zhou]{farahead2026}
Liyan Xu, Mo~Yu, Fandong Meng, and Jie Zhou.
\newblock How far ahead do {LLMs} plan? uncovering the latent horizon in
  chain-of-thought reasoning, 2026.
\newblock arXiv:2602.02103; ICML 2026.

\bibitem[Yao et~al.(2023)Yao, Yu, Zhao, Shafran, Griffiths, Cao, and
  Narasimhan]{yao2023tot}
Shunyu Yao, Dian Yu, Jeffrey Zhao, Izhak Shafran, Thomas~L. Griffiths, Yuan
  Cao, and Karthik Narasimhan.
\newblock Tree of thoughts: Deliberate problem solving with large language
  models.
\newblock In \emph{Advances in Neural Information Processing Systems}, 2023.
\newblock arXiv:2305.10601.

\bibitem[Ye and Durrett(2022)]{ye2022}
Xi~Ye and Greg Durrett.
\newblock The unreliability of explanations in few-shot prompting for textual
  reasoning.
\newblock In \emph{Advances in Neural Information Processing Systems}, 2022.
\newblock arXiv:2205.03401.

\bibitem[Zelikman et~al.(2022)Zelikman, Wu, Mu, and Goodman]{zelikman2022star}
Eric Zelikman, Yuhuai Wu, Jesse Mu, and Noah~D. Goodman.
\newblock {STaR}: Bootstrapping reasoning with reasoning.
\newblock In \emph{Advances in Neural Information Processing Systems}, 2022.
\newblock arXiv:2203.14465.

\bibitem[Zelikman et~al.(2024)Zelikman, Harik, Shao, Jayasiri, Haber, and
  Goodman]{zelikman2024quiet}
Eric Zelikman, Georges Harik, Yijia Shao, Varuna Jayasiri, Nick Haber, and
  Noah~D. Goodman.
\newblock {Quiet-STaR}: Language models can teach themselves to think before
  speaking, 2024.
\newblock arXiv:2403.09629.

\end{thebibliography}

\appendix
\section{Matched cross-comparison: design, gates, and outcome}
\label{app:mx}

\paragraph{Design.} Two models (Qwen3.5-4B, Qwen3-8B) were crossed with two tasks (DD depth-2 and MMLU,
the latter using the item pool and splits of Section~\ref{sec:b3q}). Every cell used the protocol of
Section~\ref{sec:b3q}: thinking enabled, labels extracted deterministically from the model's own rollout
($F$ the parsed final answer, $M_t$ the first candidate referenced in the next 64 tokens), decision
positions at every thinking position with $M_t$ defined and $M_t \neq F$, a full-prefix attention reader
with position~0 masked as the primary reader and an unmasked reader as a control, three seeds, and layers
at relative depths $17/28$ and $24/28$. Rules, item sets, and hashes were fixed in an internal record
before rollouts were generated; amendments are recorded there by commit.

\paragraph{Gates.} A cell receives statistics only if all of the following pass at a layer. (V1) Every
masked $M_t$ and $F$ probe trains: the median gradient norm of the reader query is at least $10^{-3}$, the
mean training loss of the final epoch is below the loss at the first step, and the final-epoch mean of the
largest attention weight is below $0.9$. (V2) Within a seed, initializations are identical across all targets, including nulls, and both readers.
(V3) The label-permutation null does not exceed the majority baseline by more than $0.05$. Driver checks on the extraction must also pass. Before a cell's main run, probe smoke runs (24, then 128
items) required the three V1 conditions for all 16 heads of each cell, including nulls and the unmasked
reader. After they failed, a full-sample diagnostic was added, with its rule fixed before it ran: every
$M_t$ head of both readers (12 per cell) had to lower a fixed-sample training loss at its best-validation
epoch and at the end, and lower validation loss at its best epoch. These two pre-main checks were not in the
original pre-specification, which treated the probe smoke run as non-decisive. The baseline for each target and layer is the larger
of the null and the majority letter on validation data, then held fixed. The primary statistic is the
difference of lifts $M_t - F$ on confirmation questions with $M_t \neq F$ at a 64-token horizon,
best-validation checkpoint, with question-level bootstrap intervals at $97.5\%$ (Bonferroni over two
layers). The model contrast on MMLU is a paired bootstrap over the confirmation questions shared by the two
MMLU cells, fixed before either cell was scored.

\begin{table}[h]
\centering
\small
\begin{tabular}{llll}
\toprule
Cell & Model $\times$ task & Outcome & Role here \\
\midrule
C1 & Qwen3.5-4B $\times$ DD & incomplete: $M_t$ probes did not train (\NMXcOneTrainOk/\NMXcOneHeads) & optimization failure \\
C2x & Qwen3.5-4B $\times$ MMLU & both readable, $F$ stronger (post-hoc) & exploratory support \\
C3 & Qwen3-8B $\times$ DD & no statistics: V1 failed at both layers & validity failure \\
C4 & Qwen3-8B $\times$ MMLU & both readable, $F$ stronger (Section~\ref{sec:b3q}) & main result (already read) \\
\bottomrule
\end{tabular}
\caption{Outcome of each cell of the attempted matched comparison. C2x is the Qwen3.5-4B MMLU cell scored
under a protocol written after the training checks that gated its main run failed; that failure is retained.}
\label{tab:mx}
\end{table}

\paragraph{C1.} Smoke and full-sample diagnostics both failed: \NMXcOneTrainOk{} of \NMXcOneHeads{}
imminent-mention probes improved training loss and \NMXcOneValOk{} improved validation loss. The cell was
closed without changing the learning rate, initialization, or any other setting. No main probe was run. For both Qwen3.5-4B cells, activation extraction ran without gradient tracking
through the reference (non-fused) implementation after the first attempt ran out of memory; this was a
post-hoc infrastructure decision, accepted only after hidden states and logits matched the original path
bit for bit on a fixed set of inputs.

\paragraph{C3.} Label validity checks passed. The main probe ran, and V1 failed at both layers because one
imminent-mention seed ended above its first-step training loss (Section~\ref{sec:mx}); V2 and V3 passed.
No accuracies were computed. A post-hoc rescoring with V1 reported but not enforced was considered and
not performed: with the imminent-mention probes near the uniform-guess loss, an accuracy computed that way
would be hard to interpret as a difference between the two targets, and it was not approved.

\paragraph{C2x.} The post-hoc protocol changes one rule: V1 is reported and not enforced. V2, V3, the
baseline rule, the checkpoint rule, seeds, and statistics are unchanged, all three seeds are kept, and each
of the 48 probes (two layers, two readers, three seeds, four targets including nulls) is recorded for
gradient flow, validation improvement over its initial loss, and attention collapse. It was written and
fixed before the C3 or C2x confirmation results were examined. Under the original V1, recomputed on the
main run, both layers pass; the earlier failures were in the smoke and diagnostic runs that gated the main run. The cell's
evidence grade is unchanged by this.
\end{document}